\documentclass{ifacconf}
\usepackage{graphicx} 
\usepackage{dblfloatfix}
\usepackage{natbib}        
\usepackage{amsmath}
\usepackage{amssymb}
\usepackage{algorithm}
\usepackage{algpseudocode}
\usepackage{array}
\usepackage{caption}
\usepackage{subcaption}
\usepackage{booktabs} 
\usepackage{float}
\usepackage{multirow}
\usepackage{colortbl}
\usepackage[x11names,table]{xcolor}

\begin{document}
\begin{frontmatter}

\title{Online Meal Detection Based on CGM Data Dynamics}
\thanks[footnoteinfo]{This paper was accepted for publication by IFAC under a Creative Commons CC BY-NC-ND license and was presented at the IFAC Diabetes Technology Conference held on May 8–9, 2025, in Valencia, Spain.}

\author[First]{Ali Tavasoli}  
\author[First]{Heman Shakeri} 

\address[First]{Center for Diabetes Technology and the School of Data Science\\ University of Virginia, Charlottesville, VA 22903 USA\\e-mails: \{at9kf, hs9hd\}@virginia.edu.}

\begin{abstract}                
We utilize dynamical modes as features derived from Continuous Glucose Monitoring (CGM) data to detect meal events. By leveraging the inherent properties of underlying dynamics, these modes capture key aspects of glucose variability, enabling the identification of patterns and anomalies associated with meal consumption. This approach not only improves the accuracy of meal detection but also enhances the interpretability of the underlying glucose dynamics. By focusing on dynamical features, our method provides a robust framework for feature extraction, facilitating generalization across diverse datasets and ensuring reliable performance in real-world applications. The proposed technique offers significant advantages over traditional approaches, improving detection accuracy, detection delay, and system robustness.
\end{abstract}


\end{frontmatter}

\section{Introduction}
The development of artificial pancreas (AP) systems has transformed type 1 diabetes (T1D) management by automating insulin delivery to maintain blood glucose within physiological ranges. However, contemporary AP systems are hybrid closed-loop and they rely on manual meal announcements, creating a critical vulnerability in their operation, e.g. subject burn-out and human errors such as missed announcements or imprecise carbohydrate estimations \citep{atlas2014closing, nimri2016closing}. Achieving a fully closed-loop (FCL) AP system requires robust automated meal detection capabilities to eliminate this reliance on manual announcements.
Although, this advancement is crucial for all individuals with T1D,  the impact is more profound for subject groups that experience the greatest challenges with hybrid closed-loop (HCL) systems. The pediatric population exemplifies this, where factors such as ongoing growth and development, communication barriers in younger children, unpredictable eating patterns, and variability in caregiver competency across different settings create unique management hurdles. By automating meal detection and insulin delivery, FCL systems have the potential to level the playing field, enabling these subjects to achieve a level of glycemic control comparable to that of other subject groups. 

Biological systems, including healthy glucose regulatory systems, are inherently homeostatic. They actively resist perturbations and strive to maintain a stable internal environment \citep{billman2020homeostasis}. However, in T1D subjects, meal intake presents a significant challenge to this stability. A meal introduces a substantial glucose influx, triggering a cascade of complex physiological responses as the body attempts to restore equilibrium \citep{mari2020mathematical}. Moreover, ingested carbohydrates are often absorbed into the bloodstream more quickly than the pharmacokinetic and pharmacodynamic profiles of insulin can counteract them \citep{chernavvsky2016use}. Hence, effective meal detection becomes paramount for enabling FCL systems, allowing for a feed-forward action (e.g. pre-meal bolusing) that complements the feedback control mechanism and helps maintain glycemic control.

We can broadly categorize existing meal detection approaches into three main classes. The first class relies on analyzing the glucose rate-of-change (ROC), employing various techniques to estimate glucose derivatives from CGM data. These methods range from simple threshold-based approaches \citep{dassau2008detection} to Kalman filtering \citep{lee2009closed}. Recent advances in this category include polynomial fitting combined with logistic regression to capture characteristic meal-response patterns \citep{moscoso2024evaluation}. 
The second class uses model-based approaches, contrasting predicted and measured glucose trajectories. These methods often utilize variants of the Bergman minimal model combined with state estimation techniques
including Unscented Kalman Filtering \citep{mahmoudi2017fault} and moving horizon estimation \citep{kolle2017meal}. 
The third class focuses on meal size estimation alongside detection, using probabilistic methods to match observed glucose patterns with potential meal shapes \citep{cameron2009probabilistic} or employing variable state dimension algorithms that alternate between meal and no-meal models \citep{xie2016variable}.

To overcome these limitations, we introduce a novel approach exploiting the principles of dynamical systems theory and utilizing Dynamic Mode Decomposition (DMD) to analyze the stability characteristics of glucose dynamics as revealed by CGM data. DMD approximates the Koopman operator, which provides a linear representation of nonlinear dynamical systems in a higher-dimensional space of observables \citep{Mezic2005, Budisic2012koopman}. 
Our central hypothesis is that unannounced meal events manifest as distinct periods of transient instability in CGM dynamics. We posit that these instabilities are detectable through real-time examination of the the system's dynamical modes and the associated eigenvalues. During periods of metabolic stability (i.e., no meals consumed), the healthy glucose regulatory system (or a feedback-controlled AP) exhibits quasi-stable dynamics. Here, DMD eigenvalues, representing the growth or decay rates of corresponding modes, typically cluster near or below unity in magnitude. However, after a meal, these eigenvalues may exceed unity, indicating temporary instability. Simultaneously, the shapes of the DMD modes capture the specific dynamic patterns associated with this meal-induced perturbation.

Our stability-based framework only uses recent CGM data and  minimizes the dependence on individual-specific parameters by focusing on universal stability characteristics. It enables real-time detection through the use of online or windowed DMD implementations \citep{Zhang2019onlineDMD}, and it provides interpretable results in the form of dynamical modes and eigenvalues, revealing insights into the system's response to meals. We demonstrate the feasibility of this approach using both simulated and real-world CGM data. Furthermore, we provide insights into the physiological interpretability of the DMD modes and eigenvalues in the context of meal responses.
The remainder of this paper is organized as follows: Section 2 presents the mathematical framework for our stability-based detection method. Section 3 details the implementation and algorithmic considerations. Section 4 presents comprehensive validation results using both simulated and real-world data, and Section 5 discusses implications and future directions. Through this work, we aim to advance the development of fully automated artificial pancreas systems and improve glucose control for individuals with T1D.

\section{Windowed DMD of CGM}
In this study, we assume only continuous glucose monitoring (CGM) data is available for the meal detection algorithm. While CGM tracks glucose fluctuations, it provides only a partial view of the glucose regulatory system, unable to directly capture key contributors like insulin, glucagon, and hepatic glucose output. This partial observability poses a significant challenge, as the CGM signal represents only the net effect of these complex physiological processes. To address this limitation, we leverage the powerful technique of delay embedding, rooted in Takens' Theorem \citep{Takens1981}. By reconstructing the system's state space from time-delayed values of the CGM data, delay embedding allows us to effectively capture the underlying dynamical behavior even without direct access to the system's internal states, such as the unobserved hormonal dynamics.

For a given CGM time series $\{\text{CGM}_k\}_{k=1}^t$, we construct the delay-embedded state vector $x(t)$ at time $t$ as follows: 
\begin{equation} 
x(t) = x_Q(t) = [\text{CGM}_{t-Q}, \dots, \text{CGM}_t]^T 
\end{equation} 
where $Q$ is the delay horizon.  At each time $t$, our goal is to analyze the local dynamics within the preceding window of length $w$ in the delay-embedded space:
\begin{equation}\label{eq:nonlDyn}
    \begin{gathered}
        x_{k+1}=F_t(x_k), \hspace{5pt} t-w\leq k\leq t
    \end{gathered}
\end{equation}
where where $t$ represents the current time step defining the end of the analysis window, while $k$ serves as the discrete time index iterating through the data points within that window (from $t-w$ to $t$). And $x_k = x(k)$ and $F_t$ represents the mapping that governs the evolution of the system within the window.

To extract the dynamical modes, we employ Dynamic Mode Decomposition (DMD) \citep{Tu2014DMD, Dawson2016, Hemati2017DMD, Zhang2019onlineDMD, Colbrook2024DMD}. DMD is a data-driven method that identifies coherent structures in time-series data by approximating the underlying dynamics with a best-fit linear operator. Within each local window, DMD seeks a matrix $A_t$ that optimally captures the dynamics described by \eqref{eq:nonlDyn} in a linear subspace: 
\begin{equation}\label{eq:linearDyn}
x_{k+1} = A_t x_k, \hspace{5pt} t-w \leq k \leq t 
\end{equation} 
The linear approximation of the underlying dynamics provided by DMD can effectively capture the dominant modes of behavior within the short time window, particularly those associated with the transient instability caused by a meal.
Crucially, the objective of this linear approximation in \eqref{eq:linearDyn} is not to perfectly model the inherently nonlinear glucose dynamics. Instead, it aims to uncover the dominant linear modes that characterize key dynamic behaviors.\ \
The underlying assumption is that meal-induced perturbations are sufficiently strong to manifest as distinct features in these dynamical modes, particularly through changes in stability characteristics. These DMD-extracted modes represent a subset of Koopman modes \citep{Mezic2005, Budisic2012koopman, colbrook2024limits}, which together provide a complete characterization of nonlinear dynamics.
 
To perform DMD in a data-driven manner, we arrange the delay-embedded CGM data within each window into snapshot pairs. Each pair consists of a state vector at a given time, $x(k)$, and its corresponding state vector one time step later, $x(k+1)$. These pairs are then arranged into matrices $\mathbf{X}_t$ and $\mathbf{Y}_t$: \begin{equation} \mathbf{X}_t = [x(t-w), \dots, x(t-1)] \in \mathbb{R}^{Q \times w} \end{equation} \begin{equation} \mathbf{Y}_t = [x(t-w+1), \dots, x(t)] \in \mathbb{R}^{Q \times w} \end{equation} Here, $\mathbf{X}_t$ contains the ``current" states within the window, and $\mathbf{Y}_t$ contains the corresponding ``future" states. Note that each matrix has $Q$ rows, corresponding to the dimension of the delay-embedded state vectors, and $w$ columns, corresponding to the number of snapshot pairs in the window. The windowed DMD algorithm then proceeds as described in Algorithm \ref{alg:dmd}).

\begin{figure}[h]
\begin{center}
\includegraphics[width=8.4cm]{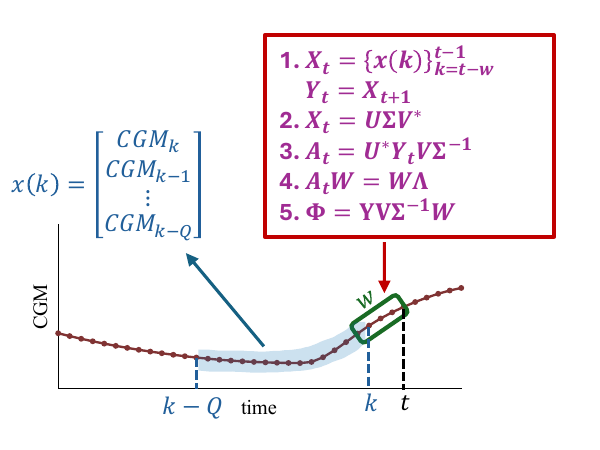}    
\caption{Schematic representation of the moving window DMD applied to CGM data in delay coordinates. The blue line represents a delay coordinate $x(k)$, and the green box indicates the sliding window of length $w$. } 
\label{fig:movingDMD}
\end{center}
\end{figure}

\begin{algorithm}
\caption{The windowed DMD algorithm.}
\label{alg:dmd}
\begin{algorithmic}[1]
\State \textbf{Input:} Snapshot data $\mathbf{X}_t \in \mathbb{R}^{Q \times w}$ and $\mathbf{Y}_t \in \mathbb{R}^{Q \times w}$, rank $r \in \mathbb{N}$.
\State Compute a $r$-truncated Singular Value Decomposition (SVD) of the data matrix $\mathbf{X}_t \approx \mathbf{U} \boldsymbol{\Sigma} \mathbf{V}^*$, 
\Statex \hspace{1.6em} $\mathbf{U} \in \mathbb{C}^{Q \times r}$, $\boldsymbol{\Sigma} \in \mathbb{R}^{r \times r}$, $\mathbf{V} \in \mathbb{R}^{w \times r}$.
\State Compute the compression $A_t = \mathbf{U}^* \mathbf{Y}_t \mathbf{V} \boldsymbol{\Sigma}^{-1} \in \mathbb{R}^{r \times r}$.
\State Compute the eigendecomposition $A_t = \mathbf{W} \boldsymbol{\Lambda}$.
\Statex \hspace{1.6em} The columns of $\mathbf{W}$ are eigenvectors and $\boldsymbol{\Lambda}$ is a diagonal matrix of eigenvalues.
\State Compute the modes $\boldsymbol{\Phi} = \mathbf{Y}_t \mathbf{V} \boldsymbol{\Sigma}^{-1} \mathbf{W}$.
\State \textbf{Output:} The eigenvalues $\text{diag}(\boldsymbol{\Lambda})\in\mathbb R^r$ and modes $\boldsymbol{\Phi} \in \mathbb{R}^{Q \times r}$.
\end{algorithmic}
\end{algorithm}
The eigenvalues in $\boldsymbol{\Lambda}$ represent the growth or decay rates of the corresponding DMD modes in $\boldsymbol{\Phi}$. The magnitude of each eigenvalue indicates the stability of the associated mode, with values greater than 1 signifying instability and values less than 1 signifying stability. The computational efficiency of the algorithm makes it suitable for online implementation.
Figure~\ref{fig:movingDMD} depicts a representation of the algorithm. 

\section{Logistic Regression of DMD modes}\label{sec:logReg}
We leverage the extracted DMD modes, that act as intrinsic dynamical features of the system,  as powerful predictors of meal events in a logistic regression model for meal detection. Specifically, we use the trace of the largest DMD eigenvalue, calculated within a 20-minute post-meal window, as a key predictor. With 5-minute CGM sampling rate, this provides four eigenvalue measurements. 
Each meal event in the training data is assigned a binary label: `1' for the presence of a meal and `0' for the absence of a meal within the specified time window. These labels, along with the corresponding DMD eigenvalue trajectories, form the training set. The logistic regression model then learns the relationship between the temporal dynamics of the dominant DMD eigenvalue and the probability of a meal occurrence. 

 \section{Results}
We evaluated our meal detection framework using the FDA-accepted UVA/PADOVA T1D simulator. We simulated 100 virtual subjects over 90 days. Meals were weight-based and scheduled at 8:00 AM (1 g/kg), 1:00 PM (0.7 g/kg), and 7:00 PM (1.2 g/kg), with random variations in timing and portions. Our analysis demonstrates that DMD features correlate with meal events. We then applied these features to develop a logistic regression model for meal prediction, as detailed in Subsection \ref{sec:logReg}.

\begin{figure*}[h!]
\begin{center}
\includegraphics[width=1\textwidth]{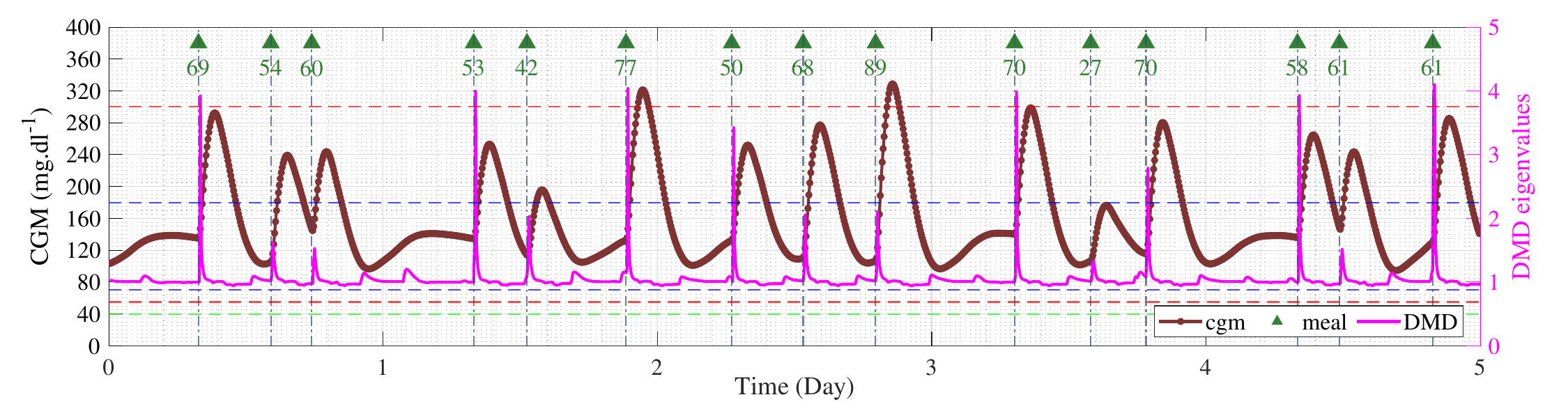}   
\caption{Correlation of meal event and DMD maximum eigenvalue, obtained by Algorithm \ref{alg:dmd}. The vertical dashed lines indicate the actual meal times.} 
\label{fig:motivation}
\end{center}
\end{figure*}
Figure \ref{fig:motivation} illustrates the real-time maximum eigenvalue of each sliding window for the CGM data, computed using Algorithm \ref{alg:dmd}. The maximum DMD eigenvalue strongly correlates with post-meal system responses, with meal-induced instability indicated by eigenvalues exceeding unity. Importantly, eigenvalue peaks precede CGM peaks, enabling earlier meal detection than traditional methods based on CGM changes or peak detection. This enables meal detection within 10-15 minutes of occurrence. Notably, this approach relies solely on CGM data with population-level tuning, obviating the need for insulin delivery information, insulin-on-board calculations, or subject-specific parameters such as total daily insulin. By leveraging dynamical modes' inherent properties, which remain relatively consistent across individuals, the method minimizes the need for personalization.

\begin{figure}[h]
\begin{center}
\includegraphics[width=0.48\columnwidth]{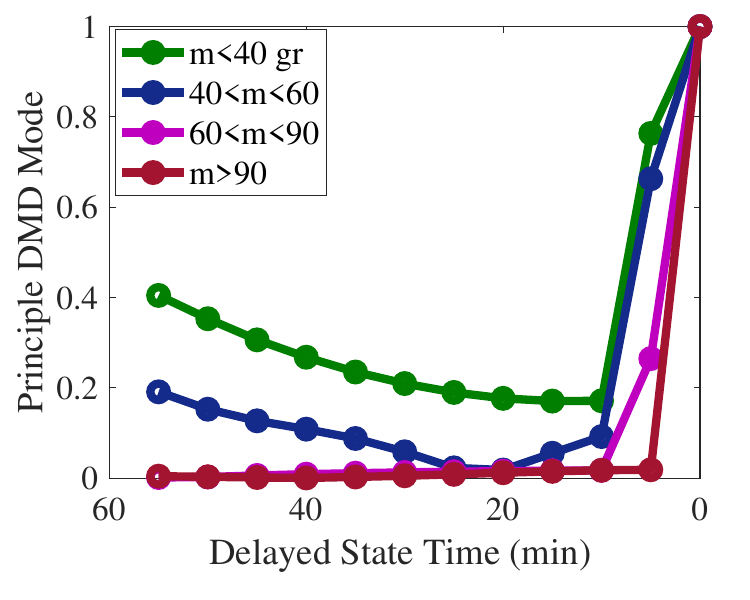}    
\includegraphics[width=0.48\columnwidth]{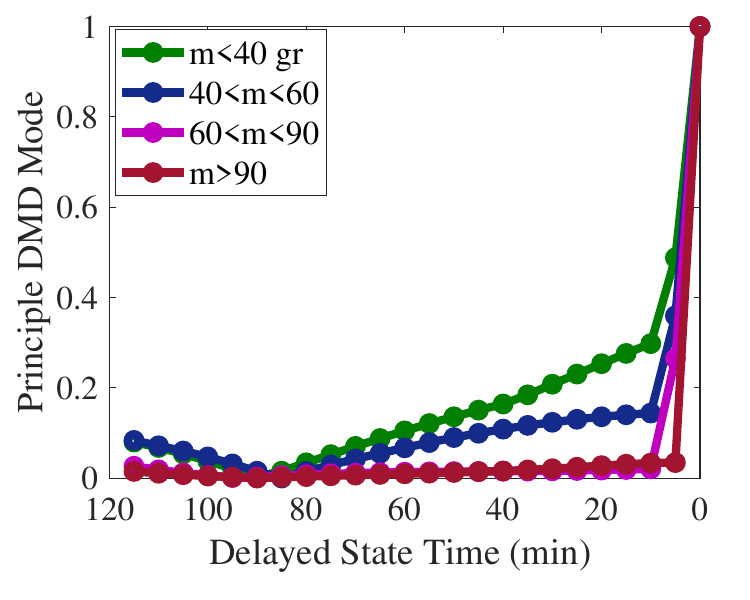} 
\caption{Principal DMD modes for different meal categories, calculated using delay horizons of $Q=60$ minutes (left) and $Q=120$ minutes (right). The shapes of the modes show a clear correlation with meal size, with larger meals exhibiting more pronounced variations.} 
\label{fig:modes}
\end{center}
\end{figure}

Figure \ref{fig:modes} illustrates the principal DMD modes, which represent the eigenfunctions associated with the maximum eigenvalues for different meal categories. A clear correlation emerges between the shapes of these DMD modes and meal sizes, where larger meals produce more pronounced and sharper variations in the corresponding modes, while smaller meals result in more subtle shape changes. This, along with our earlier findings, reveals that principal eigenvalues (Figure \ref{fig:motivation}) primarily capture meal timing, while their corresponding DMD modes (Figure \ref{fig:modes}) encode meal magnitude.

Another notable observation in Figure \ref{fig:modes} concerns the relationship between DMD modes and delay-embedded subspace selection. Comparing delays of $Q=60$ and $Q=120$ minutes reveals that modes associated with larger meals maintain relatively consistent shapes across different delay horizons, while those corresponding to smaller meals show more significant variations. This suggests that dynamical features of larger meals are more robust, persisting across both short-term and long-term time scales. In contrast, smaller meal features appear more sensitive to the choice of delay horizon, indicating their more localized temporal nature. This sensitivity might complicate detecting smaller meals. However, it also shows how variations across delay-embedded subspaces can reveal richer dynamical features. This offers valuable insights beyond meal detection into broader glucose dynamics. Hence, a multi-scale analysis enables a deeper understanding of system behavior and supports the development of more sophisticated data-driven methods in future works.

\begin{figure}[h]
\centering
    \begin{subfigure}[t]{1\columnwidth}
        \centering
\includegraphics[width=\columnwidth]{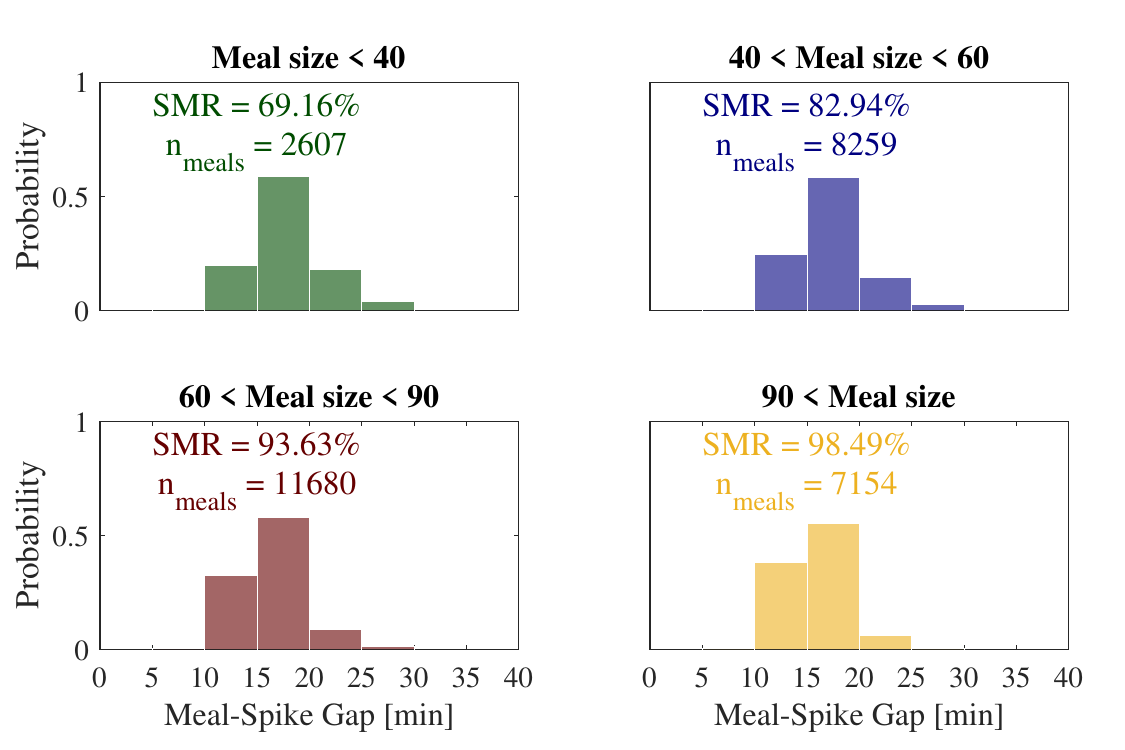}
\end{subfigure} 
\begin{subfigure}[t]{0.49\columnwidth}
        \centering
\includegraphics[width=1\columnwidth]
{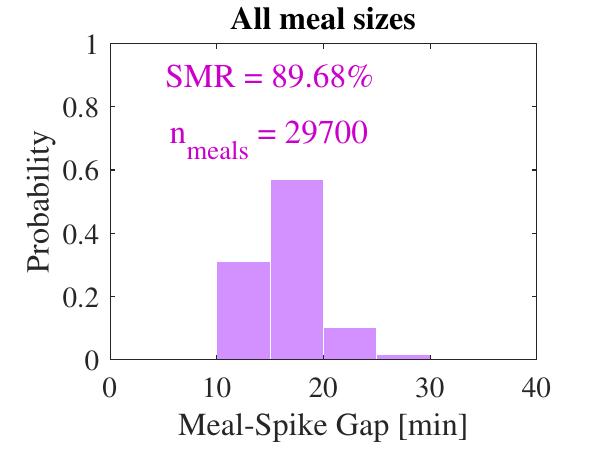}
\end{subfigure}
\caption{Meal-spike gap (meals in gram) is defined as the time interval between each meal and the first subsequent spike, detected within a 40-minute window. The spike-to-meal ratio (SMR) is defined as SMR = $\frac{n_{spikes}}{n_{meals}} \times 100$. The plots depict the time gap between each meal and the maximum spike within a 40-minute window following the meal, provided that the maximum spike exceeds a threshold of 1.2 ($\lambda_{\text{max}} > 1.2$), categorized by different meal sizes and aggregated across all meal sizes.} 
\label{fig:meal-spike}
\end{figure}

Figure~\ref{fig:meal-spike} quantifies the temporal response of DMD modes to meal intake through correlation statistics derived from 29,700 simulated meals. The analysis reveals a clear meal- size dependent pattern in detection sensitivity. Small meals (meal size  $m<$ 40 g) show the lowest correlation, with 69\% triggering a DMD spike within 30 minutes. This detection rate increases substantially for larger portions: over 82\% for medium meals (40–60 g), above 93\% for meals between 60–90 g, and exceeding 98\% for large meals (m $>$ 90 g). These findings align with our previous observation that smaller meals generate subtler variations, which can be masked by rapid dynamic or physiological fluctuations. Despite these challenges, the method demonstrates robust performance across aggregate categories, with over 89\% of meals triggering DMD spikes, typically within 10-15 minutes of intake.

\begin{figure}[h]
\centering
    \begin{subfigure}[t]{0.49\columnwidth}
        \centering
\includegraphics[width=\columnwidth]{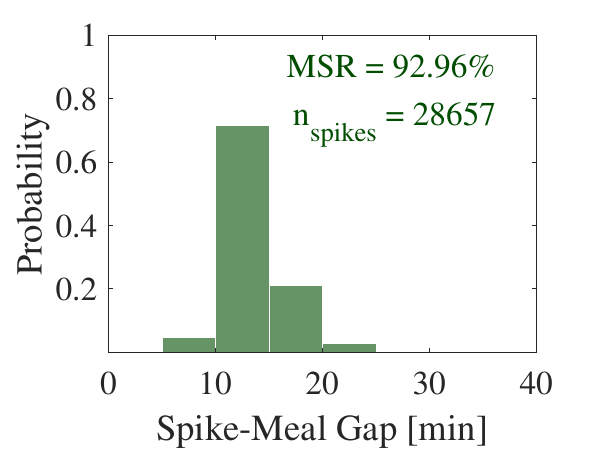}
\caption{}
\label{fig:SP}
\end{subfigure} 
\hfill
\begin{subfigure}[t]{0.49\columnwidth}
        \centering
\includegraphics[width=1\columnwidth]{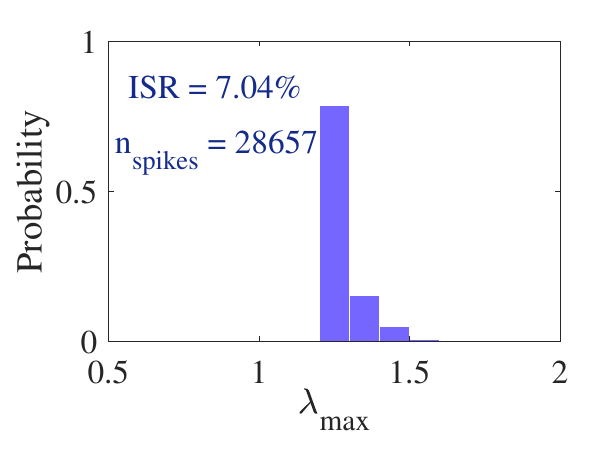}
\caption{}
\label{fig:irrelevant}
\end{subfigure}
\caption{The temporal relationship between DMD spikes and meals within a 1-hour window. (a) Shows the spike-meal gap (time interval between each spike and its preceding meal) across all meal categories, where meal-to-spike ratio (MSR) = $\frac{n_{meals}}{n_{spikes}} \times 100$. (b) Illustrates irrelevant spikes (eigenvalues $>$ 1.2 not associated with meals in the preceding hour), with irrelevant spike ratio (ISR) defined as the proportion of these spikes to total spikes.} 
\label{fig:SMgaps}
\end{figure}

Figure~\ref{fig:SMgaps} illustrates the spike-meal gap analysis not associated with meals within the previous 60 minutes. While Figure~\ref{fig:SP} shows that approximately 93\% of spikes are meal-related, the remaining 7\% represent irrelevant spikes. As shown in Figure~\ref{fig:irrelevant}, these irrelevant spikes cluster around $\lambda_{max}=1.2$, while notably, all spikes with $\lambda_{max}>1.5$ correspond to actual meals. This distribution pattern, combined with the varying detection sensitivity across meal sizes, underscores the need for a regularized logistic regression model to reliably distinguish meal-induced responses from background fluctuations in the DMD features while minimizing false positives and false negatives.

\subsection{Training on the UVA/PADOVA simulator and testing on real data} We trained our meal detection algorithm using the UVA/PADOVA simulator data and validated it against real-world CGM data from two clinical studies:

\begin{itemize} 
\item[] \underline{First study (NCT04877730):} 24 participants over two full days, comparing hybrid closed-loop therapy with UVA's Rocket artificial pancreas fully closed-loop system \citep{DCLP6}; 
\vspace{5pt}
\item[] \underline{Second study (NCT05528770):} 11 participants over two full days in hotel settings using UVA's closed-loop system \citep{moscoso2024evaluation}. 
\end{itemize}
Both studies provided comprehensive data including CGM measurements, insulin administration, rescue hypo-treatments, and staff-recorded CHO intake timing and amounts. All signals were aligned to a 5-minute time grid ($T_s = 5$), with gaps under 30 minutes filled using linear interpolation for CGM data and zeros for asynchronous signals.

For the DMD computation at each time step $t$, we used a delay horizon $Q=60$ minutes (corresponding to 12 CGM samples at $T_s=5$ min) and analyzed the preceding 15 minutes of CGM data. This 15-minute interval yields $w=3$ snapshot pairs for constructing the DMD matrices $\mathbf{X}_t$ and $\mathbf{Y}_t$ (Eq. 4-5). A rank truncation $r=3$ was applied during the SVD step (Algorithm 1) to capture the dominant dynamics. The logistic regression model was then trained on simulator data using these $r=3$ computed DMD eigenvalues at each time step as input features. We benchmark our approach against the Bolus Priming System (BPS) which estimates meal probability using CGM data from the preceding 30 minutes \citep{rocket}. The BPS algorithm classifies probabilities below 0.2 as no meal and above 0.4 as a confirmed meal, and has been extensively validated through clinical trials as a reliable method for meal detection in fully closed-loop AID systems \citep{moscoso2024evaluation}.

\begin{figure}[h]
\begin{center}
\includegraphics[width=1\columnwidth]{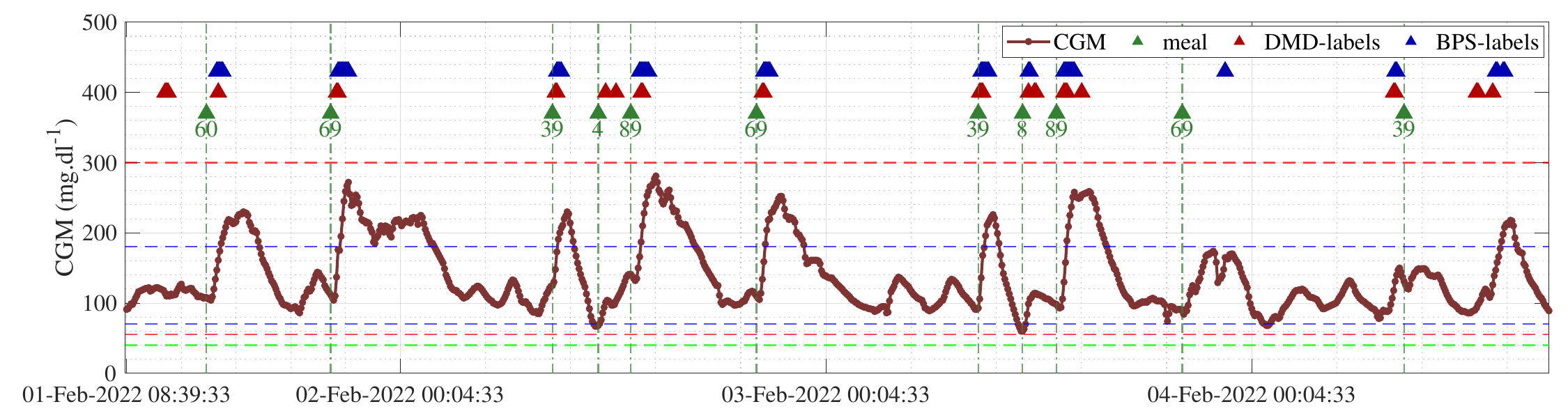}  \includegraphics[width=1\columnwidth]{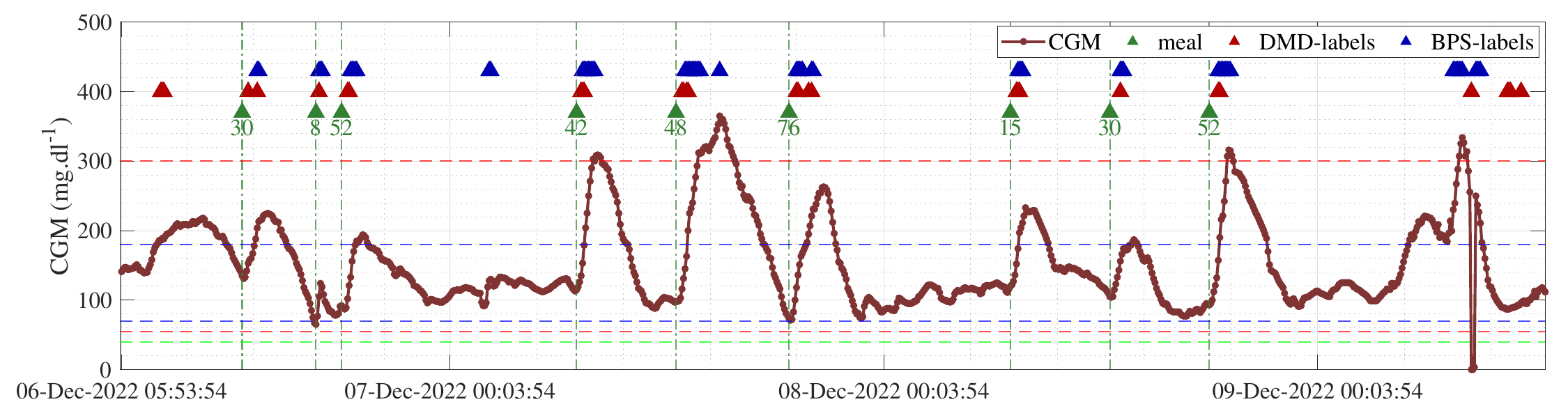}
\caption{Example test on first and second trial datasets.}
\label{fig:realData}
\end{center}
\end{figure}

\begin{figure}[h]
\centering
    \begin{subfigure}[t]{0.49\columnwidth}
        \centering
\includegraphics[width=\columnwidth]{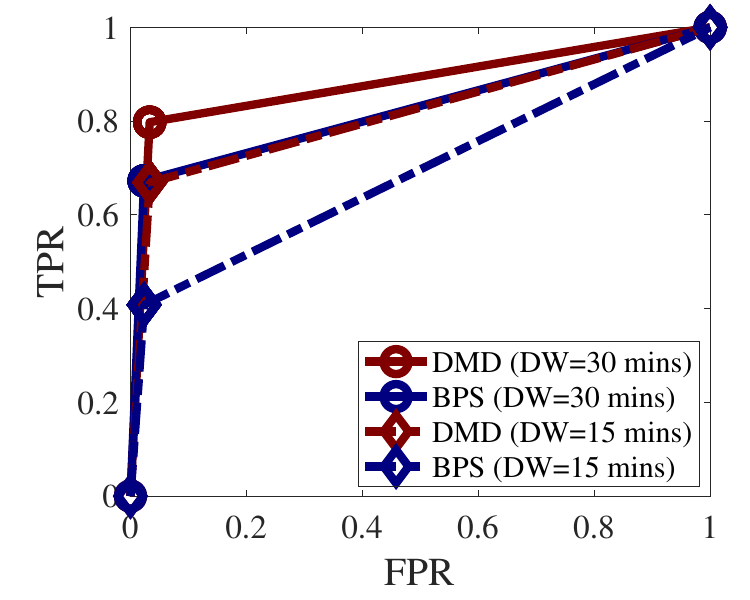}
\caption{}
\label{fig:rocDCLP6}
\end{subfigure} 
\hfill
\begin{subfigure}[t]{0.49\columnwidth}
        \centering
\includegraphics[width=1\columnwidth]{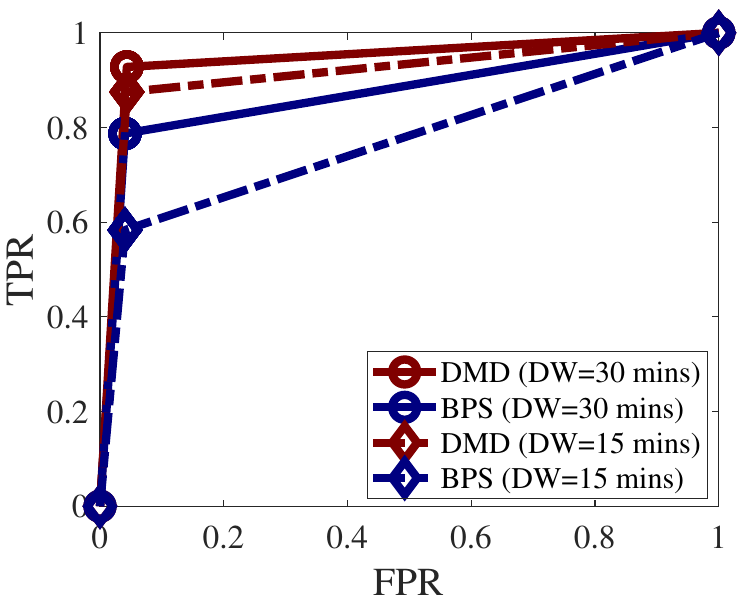}
\caption{}
\label{fig:rocBPS}
\end{subfigure}
\caption{ROC curves for (a) first, and (b) second datasets.} 
\label{fig:roc}
\end{figure}

Figure~\ref{fig:realData} demonstrates comparable high precision in meal detection performance between both algorithms across datasets. A detailed subject-level analysis suggests that errors--false positives and negatives--stem primarily from imprecise documentation of meals and hypo-treatments. The DMD approach outperforms BPS, particularly in narrower time windows (TW). As shown in Figure~\ref{fig:roc}, reducing the TW from 30 to 15 minutes decreases DMD's true positive rates (Recall) from 0.8 to 0.67 in the first dataset and from 0.93 to 0.88 in the second dataset. In contrast, BPS performance declines more substantially, with Recall dropping from 0.67 to 0.41 for the first dataset and from 0.79 to 0.58 in the second dataset. Table~\ref{tbl:metrics} provides detailed performance metrics, including Recall (equivalent to True Positive Rate or Sensitivity), False Positive Rate (FPR), and the Area Under the Receiver Operating Characteristic Curve (AUC), comparing performance across different detection time windows (TW).

\begin{table}
\centering
\caption{Meal Detection Results for Different Time Windows (TW)}
\label{tbl:metrics}
\resizebox{\columnwidth}{!}{
\begin{tabular}{@{}lc|ccc|ccc@{}}
\toprule[1pt]
\textbf{Dataset} & \textbf{Approach} & \multicolumn{3}{c|}{\textbf{TW = 30 min}} & \multicolumn{3}{c}{\textbf{TW = 15 min}} \\ 
\cmidrule(lr){3-5} \cmidrule(lr){6-8}
 &  &  \textbf{Recall} & \textbf{FPR} & \textbf{AUC} & \textbf{Recall} & \textbf{FPR} & \textbf{AUC} \\ 
\midrule
\rowcolor{red!10}
\cellcolor{white} & DMD &  0.80 & 0.03 & 0.88 & 0.67 & 0.03 & 0.82 \\ 
\rowcolor{blue!10}
\cellcolor{white} \multirow{-2}{*}{\textbf{First}} & BPS  & 0.67 & 0.02 & 0.82 & 0.41 & 0.02 & 0.69 \\ 
\rowcolor{red!30}
\cellcolor{white}  & DMD & 0.93 & 0.04 & 0.94 & 0.88 & 0.04 & 0.92 \\ 
\rowcolor{blue!30}
\cellcolor{white} \multirow{-2}{*}{\textbf{Second}} & BPS  & 0.79 & 0.04 & 0.87 & 0.58 & 0.04 & 0.77 \\ 
\bottomrule[1pt]
\end{tabular}
}
\end{table}

\begin{figure}[h]
\centering
\includegraphics[width=0.6\columnwidth]{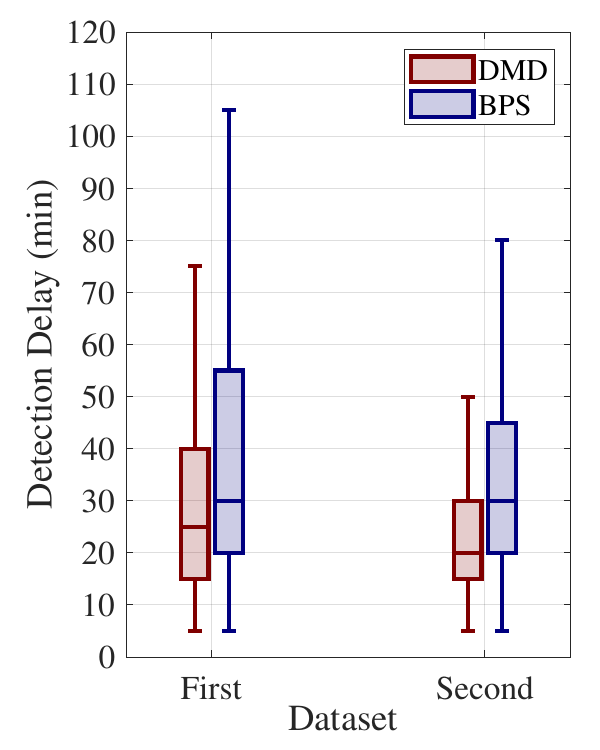}
\caption{Detection delays for different datasets.}
\label{fig:detectionDelay}
\end{figure}

Our method excels at early detection, identifying 67\% of meals in the first dataset and 88\% in the second within just 15 minutes of intake (Table~\ref{tbl:metrics}). Figure~\ref{fig:detectionDelay} provides a comprehensive analysis of detection latency, examining the time between meal consumption and the first positive detection within a 120-minute window. The DMD approach outperforms BPS with shorter median detection times: 25 minutes versus 30 minutes in the first dataset, and an even more pronounced advantage in the second dataset where DMD's median detection time reduces to 20 minutes. Moreover, DMD shows tighter detection time distributions, with smaller interquartile ranges than BPS in both datasets, highlighting its consistent performance.

\section{Conclusion}
This paper presented a real-time meal detection method based on CGM data dynamics, emphasizing that characterizing the system's underlying dynamical properties is essential for capturing inherent mechanisms that remain consistent at the population level. We introduced a windowed DMD approach in delayed dynamical subspaces to analyze CGM data, that exhibit a strong correlation between glucose system dynamics during mealtimes and DMD eigenvalues and modes.
Leveraging these dynamical features, we developed a logistic regression model trained on in-silico data from the FDA accepted UVA/Padova simulator. The model, tested on two clinical trial datasets, generalized well without needing retraining or individual fine-tuning. The algorithm outperformed the clinically validated BPS technique, achieving higher true positive rates and reduced meal detection delay.
By treating glucose monitoring as a dynamical system and analyzing its stability characteristics, rather than using traditional rate-based or pattern detection methods, our framework achieves improved accuracy across diverse datasets without requiring explicit system models. Future work will integrate this approach into fully closed-loop AID systems to assess its benefits within established control algorithms, while also expanding the use of data-driven dynamical features to address broader control objectives.

\begin{ack}
Funding Information: National Institutes of Health / National Institute for Diabetes and Digestive and Kidney Diseases (NIDDK) Grant RO1 DK 133148.
\end{ack}

\bibliography{refs} 
\end{document}